\documentclass{article}
\usepackage{caption}

\usepackage{microtype}
\usepackage{graphicx}
\usepackage{subfigure}
\usepackage{booktabs} 

\usepackage{hyperref}

\usepackage[accepted]{icml2025}

\usepackage{amsmath}
\usepackage{amssymb}
\usepackage{mathtools}
\usepackage{amsthm}
\usepackage[frozencache,cachedir=minted]{minted}

\usepackage[capitalize,noabbrev]{cleveref}

\theoremstyle{plain}

\theoremstyle{definition}

\theoremstyle{remark}

\usepackage[textsize=tiny]{todonotes}

\icmltitlerunning{To FP8 and Back Again}

\begin{document}

\twocolumn[
\icmltitle{To FP8 and Back Again: \\
Quantifying Reduced Precision Effects on LLM Training Stability}




\begin{icmlauthorlist}
\icmlauthor{Joonhyung Lee}{comp}
\icmlauthor{Jeongin Bae}{comp}
\icmlauthor{Byeongwook Kim}{comp}
\icmlauthor{Se Jung Kwon}{comp}
\icmlauthor{Dongsoo Lee}{comp}
\end{icmlauthorlist}

\icmlaffiliation{comp}{NAVER Cloud, Seongnam, Republic of Korea}


\icmlkeywords{LLM, FP8, Training, Stability, Cost}

\vskip 0.3in
]



\printAffiliationsAndNotice{}  

\begin{abstract}
The massive computational costs associated with large language model (LLM) pretraining have spurred great interest in reduced-precision floating-point representations to accelerate the process. As a result, the BrainFloat16 (BF16) precision has become the \textit{de facto} standard for LLM training, with hardware support included in recent generations of accelerators. This trend has gone even further in the latest processors, where FP8 has recently been introduced. However, prior experience with FP16, which was found to be less stable than BF16, raises concerns as to whether FP8, with even fewer bits than FP16, can be a cost-effective option for LLM training. We argue that reduced-precision training schemes must have similar training stability and hyperparameter sensitivities to their higher-precision counterparts in order to be cost-effective. However, we find that currently available methods for FP8 training are not robust enough to allow their use as economical replacements. This prompts us to investigate the stability of reduced-precision LLM training in terms of robustness across random seeds, learning rates, and datasets. To this end, we propose new evaluation techniques and a new metric for quantifying loss landscape sharpness in autoregressive language models. By simulating incremental bit reductions in floating-point representations, we analyze the relationship between representational power and training stability with the intent of aiding future research into the field.

\end{abstract}

\section{Introduction}
Conversational large language models (LLMs), such as ChatGPT \citep{openai2024gpt4}, Gemini \citep{geminiteam2024gemini1_0, geminiteam2024gemini1_5}, Claude \citep{Anthropic}, DeepSeek \citep{deepseekai2024deepseekv3technicalreport} and HyperCLOVA \citep{yoo2024hyperclova}, have captured the imagination of both academics and the public at large with their ability to communicate fluently with humans in natural language. However, these models require unprecedented amounts of computation to train, which has engendered interest in methods to improve their training efficiency.

A popular method of improving computational performance is to reduce the bit count of the floating-point representations used for training \citep{NEURIPS2018_335d3d1c, NEURIPS2020_13b91943, fp8lm}. Because reading memory is the main bottleneck in modern processors, a problem known as the “memory wall” \citep{wulf1995hitting, kim2023squeezellm}, reducing the number of bits that each floating-point number uses can accelerate the computation in proportion to the amount of memory reduced. For example, in processors that support it, computations in BrainFloat16 (BF16) \citep{DBLP:journals/corr/abs-1905-12322} can have double the maximum throughput of single precision FP32. Furthermore, the FP32 data type, the highest precision data type used in deep learning, has only half the bits of FP64, the most widely used floating-point data type in scientific computing. The current best practice for LLM training is to use BF16 for most of the LLM training computation, with some sensitive portions, such as layer normalization \citep{ba2016layer}, carried out in FP32.

As a natural extension of this development, 8-bit floating-point (FP8) \citep{NEURIPS2018_335d3d1c, NEURIPS2019_65fc9fb4, micikevicius2022fp8, fp8lm} and even 4-bit floating-point \citep{NEURIPS2020_13b91943} data formats have been proposed to accelerate training even further. However, the naïve application of FP8 to LLM training is unstable and requires additional techniques to become viable \citep{MLSYS2024_dea9b4b6}. While several methods have been proposed to stabilize training LLMs with FP8, relatively little attention has been paid to quantifying the decrease in stability compared to mixed-precision BF16 training.

Cost reduction is the motivation behind the use of FP8 and other reduced-precision training schemes as there is no intrinsic benefit in training models using smaller data types. Therefore, our concern is not whether LLM pretraining with FP8 is possible but whether it is profitable. For cost savings to be realized, the time per training step must be reduced while the number of training steps is kept to a similar number. Training stability is thus a crucial factor for cost-effective LLM training, considering that additional hyperparameter searches and restarts from training failures can outweigh any gains from raw compute acceleration.


Previous experience with training LLMs in FP16 raises further concerns. Teams that have trained LLMs have found that even when gradient scaling and other techniques are applied, the FP16 data type, which has five exponent bits, is much less stable for LLM training than BF16, which has eight exponent bits as in FP32. This raises doubts as to whether FP8, which has even fewer bits than FP16, is a practical option for real-world LLM training.

We motivate our line of inquiry with some surprising findings from experiments on the nanoGPT \citep{nanoGPT} codebase, an open-source implementation of GPT-2 pretraining, where we found that even the current best practice of mixed-precision BF16 can introduce training instabilities. When we compared BF16 and TensorFloat32 (TF32) runs, where we ran training for 5\% of the original configuration, we found that the BF16 models diverged for 18 of 188 runs, or approximately 10\% of all cases, despite using the same configurations as the default run. In contrast, no cases of loss divergence were found for the 70 TF32 models trained using different random seeds. We use the TF32 data type because NVIDIA GPUs do not offer FP32 tensor cores.

This is surprising in light of the fact that most recent LLMs are trained with mixed precision BF16 without a comparison with training on TF32, which has three additional mantissa bits. However, a loss divergence rate of approximately 10\% at only 5\% of training indicates that even standard BF16 mixed-precision training may add non-trivial instability. If even mixed-precision BF16 can cause instabilities, the effects of using even fewer bits should be investigated further. 

We make the following contributions in our work:

\begin{itemize}
    \itemsep0em
    \item We analyze the narrower hyperparameter space that emerges from reducing precision by clipping mantissa bits to simulate intermediate-bit floating-point representations and find greater instability in the model when exposed to higher learning rates or ``dirtier" data.
    \item We propose a metric for quantifying the loss landscape sharpness of models that can predict when training divergence will occur. As even the removal of mantissa bits has a destabilizing effect on LLM training, we use our metric to predict loss divergences even when the loss curve itself has not yet diverged.
\end{itemize}


To prevent misunderstanding, we also clarify on points we do \textbf{not} make. 
We do not argue that LLM pretraining with FP8 is impractical, only that there must be a deeper investigation into whether proposed methods are economical. While various techniques for FP8 training \citep{fishman2024scalingfp8trainingtrilliontoken, fp8lm, xi2025coatcompressingoptimizerstates} and even FP4 training \citep{wang2025optimizinglargelanguagemodel} have been proposed, what remains unclear is whether these methods results in a cost reduction when considering factors other than raw compute speed.

During training, a single loss divergence makes redundant the GPU time from the last checkpoint up to the idle period during the restart, possibly canceling out any gains from the accelerated computation. Also, during actual FP8 training, we cannot know what the training curve would be had BF16 been used instead. If the model requires additional steps to converge to the same point, this would further increase the cost. Even if training works reasonably well on standardized datasets and configurations, practitioners wish to know whether these findings are generalizable. We seek not to discredit any particular method but to show how a new training method can prove itself robust and trustworthy.

\begin{figure}
    \centering
    \hspace{-0.25cm}
    \includegraphics[width=0.5\linewidth]{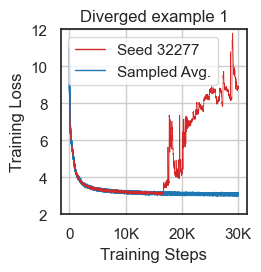}
    \includegraphics[width=0.5\linewidth]{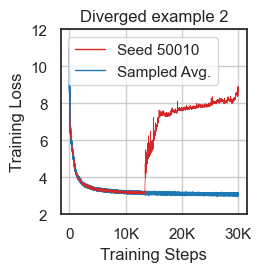}
    \caption{We show an example case of loss divergence on nanoGPT when using the same configurations as the default run except for the random seed. The blue lines indicate the average losses obtained for eight training runs that did not diverge. Of the 188 random seeds that were tested, 18 were found to diverge. As full pretraining requires over 4 days on a single node with 8 A100 GPUs, even for BF16, we perform early stopping at 30K steps, or 5\% of the original training steps, requiring approximately 4 hours for a BF16 run and 8 hours for a TF32 run per A100 node with 8 GPUs. Because we only run 5\% of the original training, we suspect that the measured divergence rate of approximately 10\% underestimates the true rate of training loss divergence.}
    \label{fig:nanoGPT}
\end{figure}

\section{Related work}

\subsection{DeepSeek V3}\label{subsection:deepseek_v3}

As we were preparing this work, DeepSeek V3 \citep{deepseekai2024deepseekv3technicalreport} has gained immense popularity as a open-source SOTA model that is capable of rivaling closed-source competitors such as ChatGPT o1. Notably, it proposes a sophisticated FP8 quantization scheme with fine-grained FP8 dynamic scaling with a group size of 128 and accumulation to FP32 for every 4 WGMMA instructions to compensate for the 14-bit accumulation of NVIDIA FP8 tensor cores. This raises a question. Had previously proposed FP8 training schemes been sufficiently stable to train LLMs cost effectively, it would not be necessary to write customized kernels. For example, in none of the previous works \citep{xi2025coatcompressingoptimizerstates, fishman2024scalingfp8trainingtrilliontoken, fp8lm, NVIDIA_TransformerEngine} on FP8 training was a scaling factor granularity of 128 proposed, with per-row scaling being the finest granularity used. Furthermore, prior to DeepSeek V3, the backward pass was implemented using E5M2, with it being the first work that we are aware of to apply E4M3 to the backward pass as well as the forward pass.

We take the unconventional FP8 implementation in the work as confirmation of our claim that previous attempts at FP8 training had not gained economical viability. Furthermore, we argue that new implementations of FP8 should undergo rigorous analysis on whether they truly are equivalent to BF16 training, and not overfit to the architectures and hyperparameters used for well-known tasks.

\subsection{Training stability}

Analyzing training instability in LLM pretraining directly is impractical due to the massive costs involved. Instead, smaller language models must be used as proxies. \citet{wortsman2024smallscale} explore the robustness of smaller language models as a proxy for LLM pretraining instability. They find that small models using a larger learning rate show similar instability patterns as larger models, such as the growth of attention layer logits and the divergence of the output logits from the log probabilities. They also explore both the causes of numerical instability in LLM training and mitigating strategies such as applying query/key (QK) normalization \citep{pmlr-v202-dehghani23a}.

\citet{keskar2017on} explore the sharpness of loss landscapes during large-batch neural network training, finding that larger batch sizes prevent the model from reaching flat regions of the loss landscape and causing performance degradation due to the inability to escape local minima. Of most significance to our work, they propose a metric for calculating loss landscape sharpness, which we adapt for LLMs as a proxy for training instability.

\citet{fishman2024scalingfp8trainingtrilliontoken} go further, discovering that Llama 7B models trained in FP8 begin to diverge after 200B tokens of training, supporting our claim that reduced-precision methods induce hidden instabilities that may emerge only later in training. They identify SwiGLU \citep{shazeer2020gluvariantsimprovetransformer} as a source of massive activations \citep{sun2024massive} and apply additional scaling factors to reduce the instability. However, it remains unclear whether the proposed method is equivalent to BF16 training or are simply stable enough for the experiments involved.

\subsection{Reduced-precision processors}
To improve throughput on computationally intensive matrix multiplication tasks, recently developed processors have been equipped with specialized hardware units such as systolic arrays for TPUs \citep{10.1145/3140659.3080246} and tensor cores \citep{nvidia_a100_whitepaper} for NVIDIA GPUs. These processors can improve throughput by an order of magnitude. For example, on the H100, the peak BF16 GEMM throughput on tensor cores is 989.4 TFLOPS, compared to 133.8 TFLOPS on CUDA cores \citep{nvidia_h100_whitepaper}.

However, the number of multiplexer circuits required for the barrel shifter of an $n$-bit floating-point unit is $n\log_2n$ \citep{barrel_shifter}, which incentivizes using smaller floating-point representations. As a result, many mixed-precision techniques perform computationally intensive matrix multiplication in BF16 while preserving sensitive portions of the model, such as the master weights and residual path activations, in FP32.

\subsection{Hybrid FP8}
The adoption of the hybrid E4M3/E5M2 formats for neural networks \citep{micikevicius2022fp8} in recent generations of processors, such as the NVIDIA H100 and the Intel Gaudi v2, has spurred interest in stable FP8 training. The hybrid FP8 format, where E4M3 is used for the forward pass for its greater resolution, and E5M2 is used for the backward pass for its greater range, was first proposed by \citet{NEURIPS2018_335d3d1c} as a means to accelerate neural networks. 

\citet{NEURIPS2019_65fc9fb4} built on this work to propose various techniques for stabilizing training, such as stochastic rounding, chunk-based accumulation, and Kahan summation during the optimizer update. However, the number of techniques that can be used in practice is limited by whether the technique in question can be applied without slowing the computation. As currently available NVIDIA GPUs, by far the accelerators with the greatest adoption, do not support these techniques natively, the overhead caused by the software-based implementations cancels out any gains from the reduced precision computations.

\subsection{MS-AMP}
Introduced in \citet{fp8lm}, MS-AMP is an automatic mixed-precision package to utilize FP8 that offers multiple optimization levels to allow for differing model sensitivities when applying reduced precision to the computations and communications of neural network training. Our experiments use the O1 optimization level of MS-AMP, which performs GEMM computations in FP8 while reducing the weights to half-precision and uses FP8 states for the weight gradients and all-reduce communications. MS-AMP offers additional optimizations for the optimizer buffer in level O2 optimization and for distributed communication in level O3 optimization, but we use only the most basic scheme so as to verify the effects of the least invasive modifications.

\section{Methods}

We seek to answer whether sub-BF16 training is competitive with standard mixed-precision BF16 training from a cost-effectiveness point of view. To be cost-effective, reduced-precision training schemes must have minimal increases in training instability and changes to hyperparameters. To better analyze the effect of reduced precision on training stability, we aim to quantify the effects of gradually reducing floating-point representations bit by bit. Hopefully, analyzing the intermediate bit representations will better illuminate the interaction between bit width and training stability.

Our intermediate-bit floating-point experiments use the TinyLlama \citep{zhang2024tinyllama} repository, which implements the widely used Llama \citep{touvron2023llama1, touvron2023llama2} architecture. TinyLlama is an open-source implementation of Llama pretraining that uses a mix of the SlimPajama \citep{cerebras2023slimpajama} and StarCoder \citep{li2023starcoder} datasets for training. It also includes performance optimizations such as Flash Attention v2 \citep{dao2023flashattention2}. We use the default learning rate of $lr=4e-4$, global batch size 512, and the same learning rate schedule as in the original code. The 120M models use a sequence length of 2048, while the 7B models use a sequence length of 4096.

\subsection{Sharpness metric}

\begin{figure}
    \centering
    \includegraphics[width=0.49\linewidth]{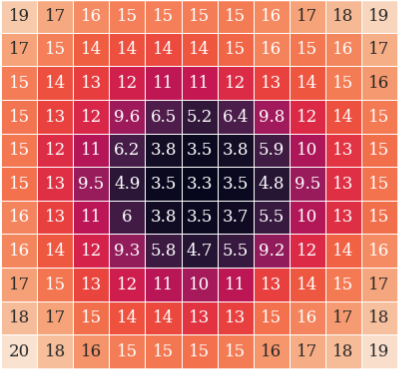}
    \includegraphics[width=0.49\linewidth]{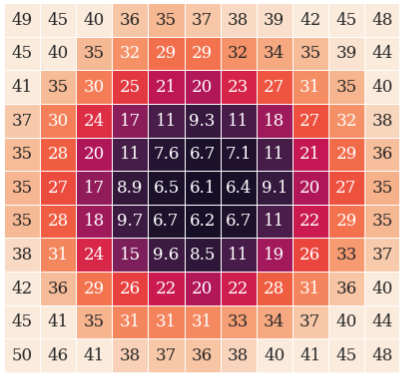}
    \caption{Loss landscape diagrams for Llama 120M E8M3 at 5K steps (left) and 10K steps (right). Even during loss divergence, the loss landscape visualized using the method in \protect\citet{visualloss} appears smooth, motivating our introduction of a new loss landscape sharpness metric. The validation loss is shown for each point of the loss landscape.}
    \label{fig:loss_landscape}
\end{figure}

\begin{figure}
\begin{minted}{python}
def forward(x, w):
    # x: input, w: weight, out: output
    save_for_backward(x, w)
    masked_x = reduce_precision(x)
    masked_w = reduce_precision(w)
    out = F.linear(masked_x, masked_w)
    masked_out = reduce_precision(out)
    return masked_out
\end{minted}
\captionof{figure}{PyTorch-like pseudocode for the forward pass.}
\label{alg:pseudocode}
\end{figure}

To better investigate the model state when loss divergence occurs, we attempted to visualize the loss landscape of the Llama models with the method proposed by \citet{visualloss}. However, as shown in Figure \ref{fig:loss_landscape}, we found that even when the model is clearly in the process of loss divergence, the generated visualizations remain smooth. We emphasize that we do not seek the loss landscape sharpness \textit{per se}, but a proxy to provide a quantitative measurement of the underlying training instability.

Because of this, we propose an alternative loss landscape sharpness metric that is more suitable for autoregressive models, based on the one proposed in \citet{keskar2017on}. We empirically confirm that it is a useful indicator of training instability in the following sections. The main difference between the original metric and our version is that we use the logit of the last token instead of the model input for the calculation. This is because adding noise to the embeddings in a language model has different implications compared to adding noise to input images in a vision model. We also do not apply random projection as in \citet{keskar2017on} to reduce the stochasticity of the measurement.

Instead of searching the input space as in \citet{keskar2017on}, we apply the search algorithm to the logit space of the last token. Searching the logit space has the additional advantage that the forward pass of the model need only be performed once for each measurement, significantly reducing the computational cost. The logit of the last token was chosen because it is not computationally feasible to optimize for the entire output space. Also, due to the autoregressive character of decoder-only Transformer models \citep{NIPS2017_3f5ee243, NEURIPS2020_1457c0d6}, the last token is the only one to receive inputs from all other tokens.

\paragraph{Definition} Let $\boldsymbol{y} \in \mathbb{R}^{s\times v}$ be the output logit for an autoregressive model of sequence length $s$ and vocabulary size $v$. Then, for $\boldsymbol{y}_i$, the output logit at sequence position $i\in\{1,2,...,s\}$, and one-vector $\mathbf{1}_v \in \mathbb{R}^v$,  we define a constraint set $\mathcal{C}_\epsilon$ at $i=s$ such that

\begin{equation}
    \mathcal{C}_\epsilon \in \{\boldsymbol{z}_s \in \mathbb{R}^v: -\epsilon(|\boldsymbol{y}_s|+\mathbf{1}_v) \le \boldsymbol{z}_s \le \epsilon(|\boldsymbol{y}_s|+\mathbf{1}_v) \}.
\end{equation}

Given $\boldsymbol{y}_s \in \mathbb{R}^v, \epsilon>0$, and noise vector $z_s$, the loss landscape sharpness $\phi_\epsilon$ for loss function $f$ can be defined as

\begin{equation}
\label{eq:sharpness}
    \phi_\epsilon:=\frac{max_{\boldsymbol{z}_s\in\mathcal{C}_\epsilon}f(\boldsymbol{y}_s+\boldsymbol{z}_s)-f(\boldsymbol{y}_s)}{1+f(\boldsymbol{y}_s)} \times 100.
\end{equation}

 The proposed metric can best be thought of as the relative magnitude of the largest loss spike on the logit within the provided bounds. The bounds are set to be the logit magnitudes plus one multiplied by $\epsilon$. The largest spike in the vicinity of the logits is found using the L-BFGS-B algorithm \citep{Liu1989-ux}, using the SciPy \citep{2020SciPy-NMeth} implementation with the output logit set as the starting point of the search. We set $\epsilon=5e\mathrm{-4}$ for all our experiments following \citet{keskar2017on}. However, a hyperparameter sweep on $\epsilon$ shows that the general trend is unaffected by the value of $\epsilon$, only the sharpness value magnitudes. The results are shown in Table \ref{tab:epsilons} of the Appendix.

\subsection{Masking}

Our experiments use a simplified method of reducing floating-point precision to achieve reasonable throughput. To simulate removing exponent bits, we threshold the values to the minimum and maximum absolute values possible with the given number of exponent bits. Figure \ref{fig:exp_mask} depicts the exponent masking process. A bitmask is applied to remove the unrepresentable mantissa bits. The resulting method is an imperfect approximation of reducing floating-point bit count. However, it has the advantage of being fast, causing at most a doubling of the time per training step.

\begin{figure*}
  \centering
  \includegraphics[width=\linewidth]{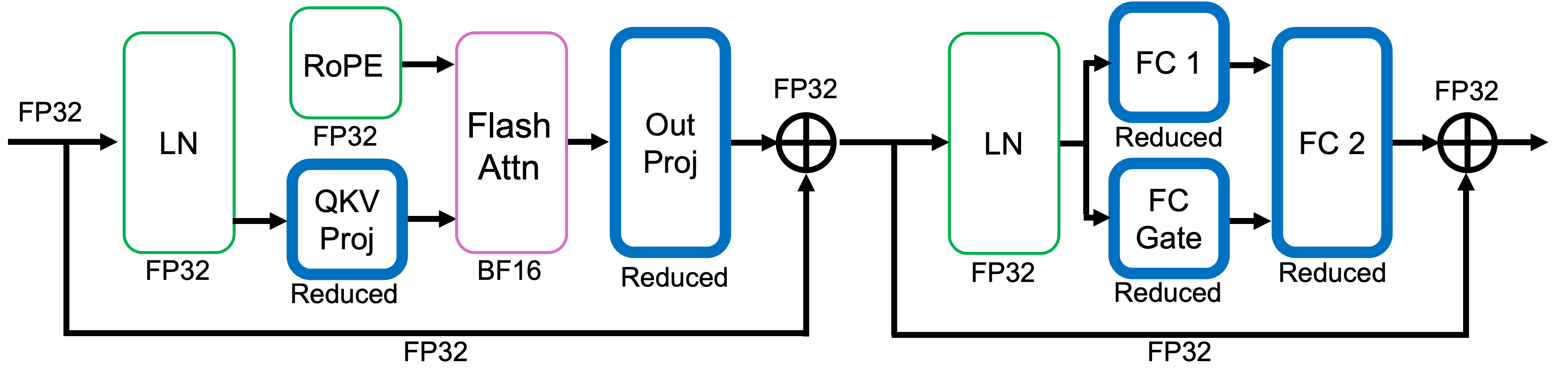}
  \caption{
  Diagram showing the precisions used in a Llama decoder block (best seen in color).
  The activations in the path of the residual connection are kept in FP32, as are the model weights and embeddings.
  The LayerNorm and RoPE \protect\citep{RoFormer} layers use FP32 internally for their computations. The Flash Attention kernel uses BF16 with no reduction in precision. All other layers use reduced-precision matrix multiplication that emulates low-precision computation with a high-precision accumulator.
  }
  \label{fig:precision_diagram}
\end{figure*}

We apply reduced precision operations only on the matrix multiplication computations of the model, excluding the attention computation, which uses the Flash Attention v2 kernel. Following existing FP8 libraries such as TransformerEngine \citep{NVIDIA_TransformerEngine}, we separate the effects of reducing the computation's precision from reducing the data's precision in storage. As a result, the activations and model weights are kept at their original precision while the inputs and outputs of matrix multiplication are dynamically masked to emulate reduced precision computation with a high-precision accumulator. All states are kept in their original precision, and all operations other than matrix multiplication are performed in their original precision. In Figure \ref{fig:precision_diagram}, we include a diagram indicating the precision of the states and computations in a Llama decoder block.

\section{Results}
\subsection{MS-AMP experiments}
\label{subsection:msamp}

\begin{figure}
\centering
\includegraphics[width=\linewidth]{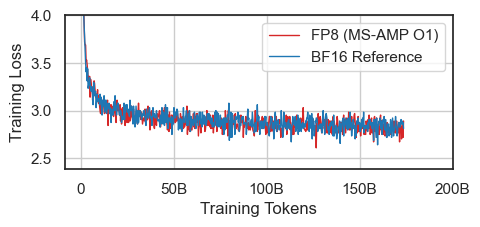}
\captionof{figure}{
Comparison between MS-AMP FP8 (O1) and BF16 training on a subsample of the FineWeb Edu dataset using a Llama 120M model.
}
\label{fig:msamp_fwe}
\end{figure}

\begin{figure}
    \centering
    \includegraphics[width=0.7\linewidth]{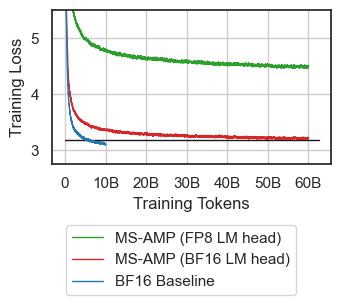}
    \caption{Training losses for GPT-2 training compared using exponential moving averages to better visualize the general trends. The blue curve indicates the training loss for the baseline BF16 training, while the red curve indicates MS-AMP level O1 training with the LM head excluded from FP8 quantization. The green curve shows the training loss for when the LM head included in the FP8 quantization.}
    \label{fig:msamp_lm_head}
\end{figure}

We first analyze the effect of real-world FP8 training by applying MS-AMP \citep{fp8lm} v0.4.0 level O1. All experiments are run on an H100 node to ensure hardware availability of FP8 with 8 GPUs. Two sets of experiments are performed. In the first, shown in Figure \ref{fig:msamp_fwe}, we train a Llama 120M model on a subsample of the FineWeb Edu \citep{penedo2024finewebdatasetsdecantingweb} dataset. In the second, shown in Figure \ref{fig:msamp_lm_head}, we train a GPT-2 124M model from the nanoGPT \citep{nanoGPT} codebase using the OpenWebText \citep{Gokaslan2019OpenWeb} dataset.

From the two experiments, we come to the following conclusions. In Figure \ref{fig:msamp_fwe}, where we train a Llama model on FineWeb Edu, a well-known ``clean" dataset, we do not observe a difference between the convergence rates of MS-AMP O1 models and BF16 reference models. However, in Figure \ref{fig:msamp_lm_head}, we see a divergence between the MS-AMP O1 results and the BF16 baseline that does not close, even after applying six times the training iterations for the MS-AMP model at 120K training steps. The differences are much greater when the LM head is included in the quantization, indicating that parts of the model are highly sensitive to the quantization, requiring further investigation into model architecture choices.

These results indicate that the FP8 training scheme in MS-AMP may not converge to the same loss as BF16 training or requires more training steps, depending on factors such as data quality. This strengthens our case that FP8 training may introduce hidden instabilities that are not evident until stress tested against circumstances that were not considered in the original works proposing them.

LLM pretraining datasets in production environments, especially for customized use-cases, are usually much ``dirtier" than those of popular open-source datasets. For example, extensive data filtering may not be an option for low-resource languages or for subfields such as medical or legal data. Also, for newer domains such as video or robotic motion, well-established metrics of data quality do not yet exist. Therefore, the finding that a method of FP8 training may work well only on ``clean" data would be a consequential one for LLM training in production.

\subsection{Bit reduction experiments}

We first attempt to identify the points where training instability becomes visible. The emulated reduced-precision representations are denoted using the number of exponent and explicit mantissa bits used. For example, standard BF16 is referred to as E8M7, while a floating-point number with its exponent clamped to seven bits and mantissa clamped to six bits is referred to as E7M6. 

\begin{figure}
    \centering
    \hspace{-0.25cm}
    \includegraphics[width=0.5\linewidth]{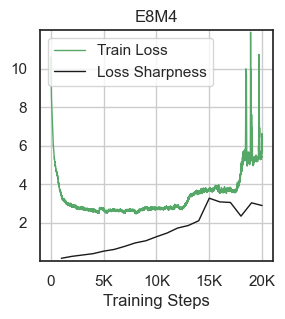}
    \includegraphics[width=0.51\linewidth]{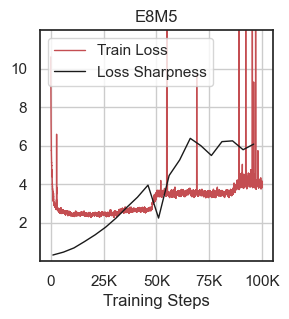}
    \caption{TinyLlama 120M models trained until loss divergence. E8M4 and E8M5 models are trained for 20K and 100K steps, respectively. The dotted black line in each figure indicates the loss landscape sharpness of the model.}
    \label{fig:tinyllama120m_e8m345}
\end{figure}

We find that removing even a single exponent bit prevents training altogether, resulting in the model failing to progress with any learning using E7M7, confirming previous findings \citep{8877427} that neural network training is more sensitive to exponent bits than mantissa bits. To analyze the cause, we conduct an ablation on the clamping mechanism by either removing only the inner or outer exponent range, as depicted in Figure \ref{fig:exp_mask} in the Appendix. We find that models with only their inner exponent ranges clamped train normally while models with only their outer exponent ranges clamped do not, indicating that the inability to represent large values is the cause of failure for E7M7. We therefore investigate the effects of removing mantissa bits for the remainder of our experiments.

\subsection{Loss landscape sharpness}

To further uncover the relationship between bit width and training robustness, we use Equation \ref{eq:sharpness} to quantify the degree of training instability increase by measuring the loss landscape instability of Llama models. In Figure \ref{fig:tinyllama120m_e8m345}, we show Llama 120M models trained until their training losses diverge, as well as plotting the loss landscape sharpness values of the models in E8M4 and E8M5. Training required approximately 2 hours per 10,000 training steps on a node with 8 A100 GPUs. 

Although the points of divergence are different for each model, we can see a general trend of increasing sharpness until the model diverges sharply, after which it cannot revert to its original training trajectory. This pattern is visible despite the large differences in training steps.

\begin{figure}
\includegraphics[width=\linewidth]{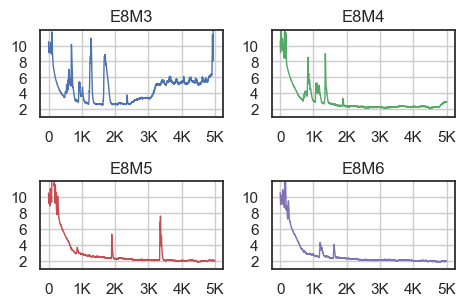}
\caption{Llama 7B model training loss curves for different mantissa bits.
The x-axis shows training steps, while the y-axis shows the training loss.}
\label{fig:llama_7b_runs}
\hfill
\end{figure}

To verify that similar behavior is reproducible in larger models, we compare the training losses of Llama 7B models trained for 5,000 steps in Figure \ref{fig:llama_7b_runs} and show the measured sharpness values for $\epsilon=5e\mathrm{-4}$ in Table \ref{tab:llama_7b}. Results for other $\epsilon$ values are included in Table \ref{tab:epsilons} of the Appendix and show a similar pattern. We apply early stopping at 5,000 training steps because training a Llama 7B model for 5K steps requires approximately one week on a single node with 8 A100 GPUs. These experiments show that loss divergence is visible in the E8M3 and E8M4 models, while it has yet to emerge in the E8M5 model. However, from Table \ref{tab:llama_7b}, we can see that the loss landscape sharpness continues to increase for the E8M5 model, even though no signs of instability are yet visible in the training loss.

\begin{table}
\centering
\small
\begin{tabular}{cccccc}
\toprule
Steps &  E8M3 &  E8M4 &  E8M5 &  E8M6 &  E8M7 \\ \midrule
   1K & 0.209 & 0.205 & 0.191 & 0.191 & 0.192 \\
   2K & 0.488 & 0.363 & 0.265 & 0.221 & 0.200 \\
   3K & 1.306 & 0.734 & 0.352 & 0.229 & 0.200 \\
   4K & 2.006 & 1.125 & 0.475 & 0.237 & 0.207 \\
   5K & 1.927 & 1.439 & 0.628 & 0.248 & 0.215 \\ \bottomrule \\
\end{tabular}
\caption{Loss landscape sharpness values at $\epsilon=5e\mathrm{-4}$ for Llama v2 7B models trained with TinyLlama for 5,000 steps in Figure \ref{fig:llama_7b_runs}. Training used a global batch size of 512 and a sequence length of 4096.}
\label{tab:llama_7b}
\end{table}

The E8M3 and E8M4 models show much higher sharpness values, and both diverge early in training. In contrast, there is only a gradual increase in the loss-landscape sharpness for the E8M7 runs. Figures \ref{fig:tinyllama120m_e8m345} and \ref{fig:llama_7b_runs} show that models gradually increase in sharpness until a threshold level is reached. However, the exact threshold may differ depending on the configurations. These results suggest that models with fewer mantissa bits enter regions of ever greater instability during training, even when these instabilities are not visible in the loss curve. We believe that, in the future, such analysis of loss landscape sharpness can be used to identify when the model is at risk of training loss divergence.

To verify that our proposed loss landscape sharpness metric tracks the training instability when it decreases as well as when it increases, we show the loss landscape sharpness trend for the E8M3 model when training is reverted to BF16 in Table \ref{tab:sharpness} of the Appendix. Supporting our claim that the loss landscape sharpness is an indicator of training instability, the sharpness value continues to decrease until it nears the level of BF16 training as shown in Table \ref{tab:llama_7b}.

\subsection{Robustness to learning rate changes}

\begin{figure}
    \centering
    \includegraphics[width=0.9\linewidth]{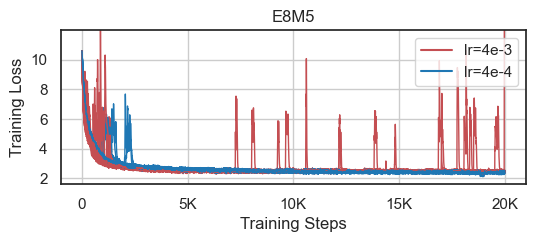}
    \includegraphics[width=0.9\linewidth]{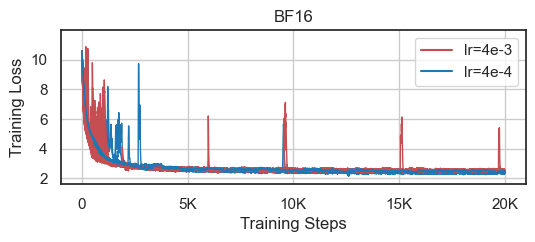}
    \caption{Comparison between Llama 120M models trained using E8M5 masked training (left) and standard BF16 training (right) for $lr=4e\mathrm{-4}$ (the default learning rate) and  $lr=4e\mathrm{-3}$. Using 18 random seeds per configuration, the E8M5 runs show more frequent loss spikes, especially at the higher learning rate, indicating greater training instability.}
    \label{fig:llama120m_20k_lr}
\end{figure}

We further attempt to identify hidden instability in E8M5, which did not diverge during the initial training stages in Figure \ref{fig:tinyllama120m_e8m345}. Inspired by \citet{wortsman2024smallscale}, we analyze the robustness of Llama 120M models to changes in the learning rate by comparing training at BF16 with that for E8M5. As seen in Figure \ref{fig:llama120m_20k_lr}, the E8M5 training runs have more frequent loss spikes during training, especially when the learning rate is increased to $4e\mathrm{-3}$. Although no cases of loss divergence were found, we believe that the higher frequency of loss spikes indicates greater sharpness of the loss landscape, supporting our claim that training is more unstable for E8M5 even before loss divergence occurs.

\section{Discussion}
This work proposes quantitative evaluations and analyses of training instabilities when reducing floating-point precision. Our experiments have shown that MS-AMP, an existing open-source implementation of FP8 training, does not offer sufficient robustness to allow its cost-effective use. By emulating lower-precision mantissa bits in matrix multiplication, we further analyze the increasing instabilities arising from lower precision in LLM training.

Again, we clarify that we are not arguing that FP8 training is not viable. Indeed, we discuss the unique characteristics of DeepSeek V3 FP8 training in Section \ref{subsection:deepseek_v3} and reproduce stable training using FP8 using the MS-AMP library in Section \ref{subsection:msamp}. The issue is that FP8 training causes a narrowing of the hyperparameter space where LLM training can occur stably and with equivalent performance as mixed-precision BF16 training. Because of this narrower hyperparameter space, more resources must be expended on identifying and honing techniques for preventing training collapse. Worse, there is simply no way of knowing if the FP8 training is performing competitively as BF16 without implementing a BF16 training run for comparison, which would negate the purpose of using FP8 for training in the first place. Even if FP8 training is practical for carefully selected hyperparameters under specific conditions, we assert that the costs of finding such conditions and the risks involved in training a less stable model outweigh the benefits of using FP8 for accelerated computation. Because of this risk, we propose methods to evaluate the stability and robustness of reduced-precision training, which is vital for FP8 or other reduced-precision training schemes to be viable for real-world LLM training. 

From our experiments, several methods naturally suggest themselves as possible stabilization techniques. First, the initial stages of training could be conducted in higher precision, similar to how smaller batch sizes may be used during the initial stages of training as in \citet{keskar2017on}. Increasing the precision when the loss landscape becomes too sharp may also provide a tradeoff between training speed and stability. Second, the more sensitive layers may be kept at high precision, while only the less sensitive layers are computed with reduced precision. For example, in Figure \ref{fig:e7m7_lm_head_bf16}, we found that removing masking from the LM head of a Llama model was sufficient to enable E7M7 training to progress, although the resulting model was unstable. For the instability in GPT models shown in Figure \ref{fig:nanoGPT}, we found that increasing the precision of the first two decoder blocks to TF32 was sufficient to prevent loss divergence. However, as such compensatory techniques depend on the model architecture, training data, and other aspects of the training environment, we leave their investigation to future work.

\section{Limitations}
A limitation of this work is that it focuses on the initial stages of pre-training when many instabilities are known to arise only later in training \citep{fishman2024scalingfp8trainingtrilliontoken, Bekman}. For example, \citet{wortsman2024smallscale} show that the logits of the outputs diverge from zero only at the later stages of training, and \citet{fishman2024scalingfp8trainingtrilliontoken} show that correlations between the SwiGLU layers only begin to increase after 200B tokens. To this, we argue that our studies likely underestimate the instabilities that FP8 or other reduced precision training schemes will face, further strengthening our argument that currently available FP8 training methods are too unstable to be profitably utilized for LLM training in their current form and that much more extensive evaluations are required to stress-test them.

Second, despite finding that the exponent bits are of greater importance to LLM training than mantissa bits, we were unable to experiment by increasing the number of exponent bits. This was because matrix multiplication in FP64 is over an order of magnitude slower than BF16 on A100 and H100 GPUs when using tensor cores. Experiments using representations such as E11M4, created by removing 48 mantissa bits from FP64, may be illuminating, but we found it impractical to train models with a greater number of exponent bits due to hardware limitations.

Finally, our experiments are limited in that they only the training loss is used as an evaluation metric
instead of real-world natural language tasks such as MMLU \citep{hendryckstest2021} scores. However, while lower perplexity does not guarantee superior performance on downstream tasks, we believe that training loss divergence is sufficient to indicate training failure.

\section{Conclusion}
We demonstrate that the training stability of LLMs decreases incrementally with the reduction of floating-point bit widths used for training models of up to 7B parameters. Using our proposed loss landscape sharpness metric, we measure the gradual narrowing of the hyperparameter space where stable training is possible. 

In future works proposing reduced-precision training, we hope that our work inspires analyses of the robustness of newly proposed training methods not only on the well trodden paths with well-established guidelines, but also on less explored conditions where many of the initial assumptions may not hold. The massive power and capital consumption of LLM training means that much is at stake on this issue.

\bibliography{references}
\bibliographystyle{icml2025}

\newpage
\appendix
\onecolumn
\begin{figure}
  \centering
  \includegraphics[width=0.4\linewidth]{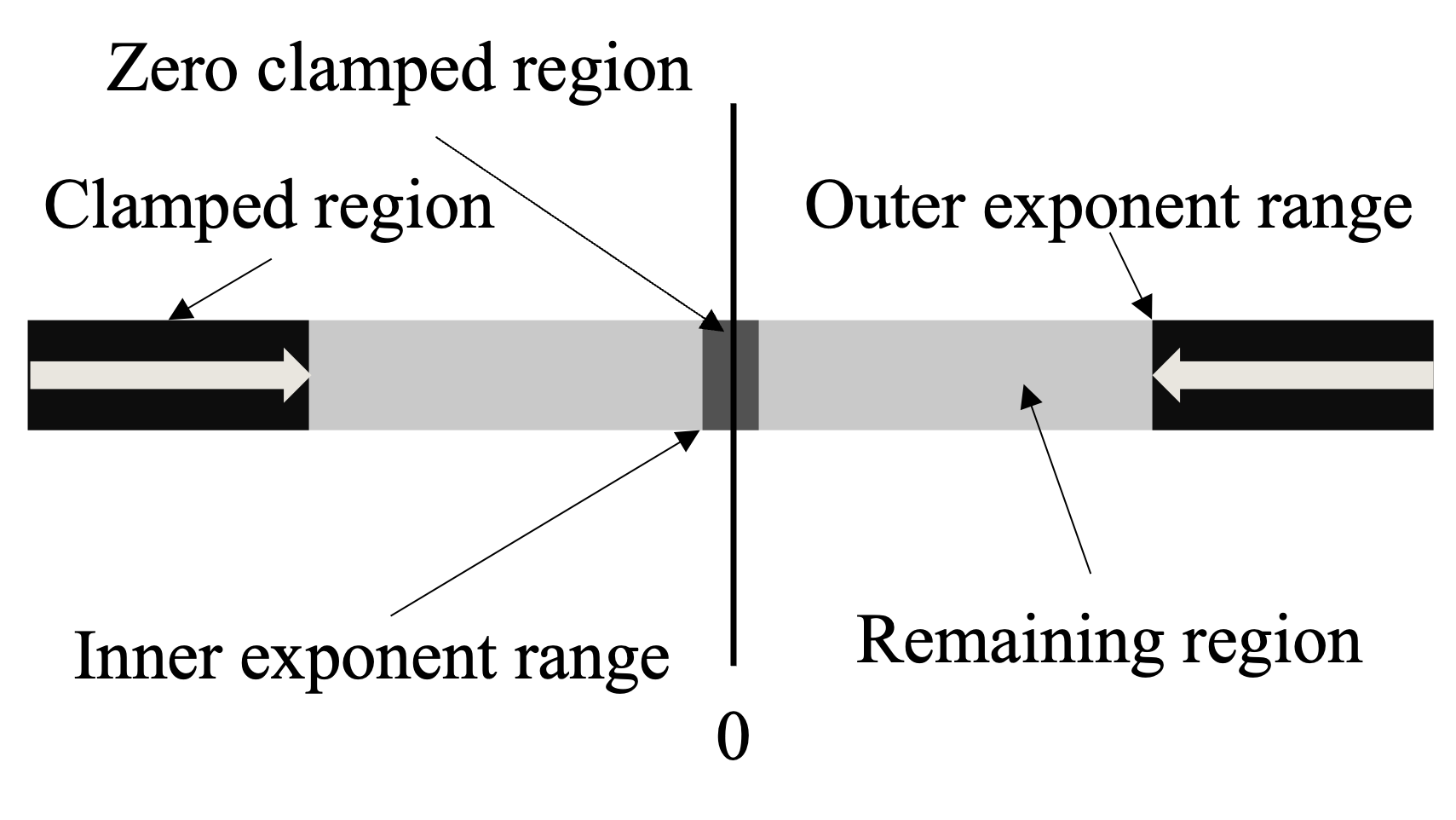}
  \caption{Exponent masking by clamping values not expressible with the given number of exponent bits.}
  \label{fig:exp_mask}
\end{figure}

\begin{figure}[H]
    \centering
    \includegraphics[width=0.8\linewidth]{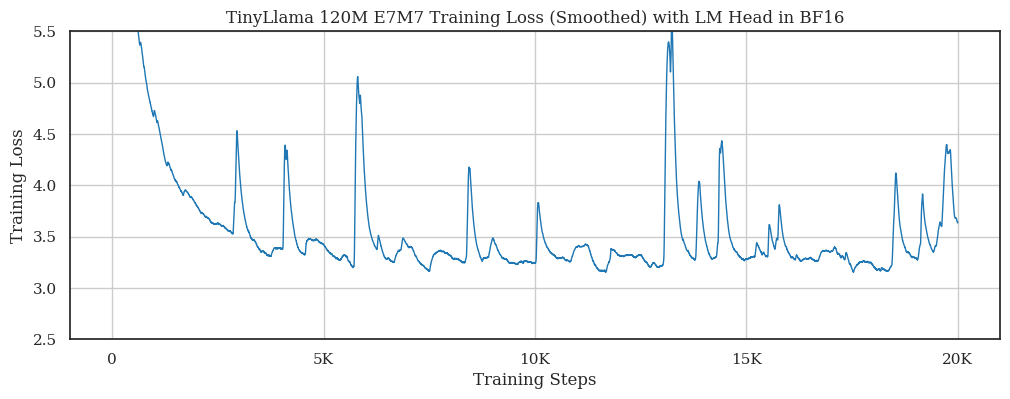}
    \caption{The training loss of a TinyLlama 120M model with clipped exponent at E7M7, excluding the LM head. The training loss is smoothed using exponential moving averages for better visualization. The results show that the exponent clipped models remain unstable when models in Figure \ref{fig:llama120m_20k_lr} have stabilized after the same number of training steps.}
    \label{fig:e7m7_lm_head_bf16}
\end{figure}

\begin{table}[ht]
\captionof{table}{Robustness of the sharpness metric to $\epsilon$. We have found empirically that the loss landscape sharpness metric is robust to the choice of $\epsilon$. Below, we include a table with sharpness values for a wide range of $\epsilon$ values for a Llama 7B model trained for 5,000 steps. We use a different checkpoint from the one in Table \ref{tab:llama_7b} of the paper to demonstrate reproducibility.
}
\begin{tabular}{ccrrrrr}
$\epsilon$ & Precision & 1K & 2K & 3K  & 4K    & 5K    \\
\hline \\
5.00E-05  & E8M3 & 0.02 & 0.06 & 0.18  & 0.19  & 0.17  \\
          & E8M4 & 0.02 & 0.04 & 0.08  & 0.13  & 0.17  \\
          & E8M5 & 0.02 & 0.03 & 0.04  & 0.05  & 0.07  \\
          & E8M6 & 0.02 & 0.02 & 0.02  & 0.02  & 0.03  \\
          & E8M7 & 0.02 & 0.02 & 0.02  & 0.02  & 0.02  \\
          &      &      &      &       &       &       \\
1.00E-04  & E8M3 & 0.04 & 0.11 & 0.36  & 0.38  & 0.33  \\
          & E8M4 & 0.04 & 0.08 & 0.16  & 0.26  & 0.34  \\
          & E8M5 & 0.04 & 0.05 & 0.07  & 0.10  & 0.14  \\
          & E8M6 & 0.04 & 0.04 & 0.04  & 0.05  & 0.05  \\
          & E8M7 & 0.03 & 0.04 & 0.04  & 0.04  & 0.04  \\
          &      &      &      &       &       &       \\
5.00E-04  & E8M3 & 0.19 & 0.51 & 1.70  & 1.80  & 1.49  \\
          & E8M4 & 0.18 & 0.37 & 0.74  & 1.22  & 1.54  \\
          & E8M5 & 0.18 & 0.25 & 0.34  & 0.48  & 0.64  \\
          & E8M6 & 0.18 & 0.21 & 0.21  & 0.23  & 0.25  \\
          & E8M7 & 0.16 & 0.18 & 0.17  & 0.19  & 0.19  \\
          &      &      &      &       &       &       \\
1.00E-03  & E8M3 & 0.38 & 0.98 & 3.31  & 3.59  & 2.81  \\
          & E8M4 & 0.36 & 0.71 & 1.42  & 2.32  & 2.86  \\
          & E8M5 & 0.34 & 0.49 & 0.66  & 0.92  & 1.21  \\
          & E8M6 & 0.34 & 0.41 & 0.40  & 0.44  & 0.48  \\
          & E8M7 & 0.30 & 0.35 & 0.34  & 0.37  & 0.38  \\
          &      &      &      &       &       &       \\
5.00E-03  & E8M3 & 1.73 & 4.47 & 13.64 & 11.64 & 9.58  \\
          & E8M4 & 1.61 & 3.26 & 6.43  & 9.87  & 11.39 \\
          & E8M5 & 1.55 & 2.25 & 2.92  & 4.10  & 5.28  \\
          & E8M6 & 1.55 & 1.87 & 1.85  & 2.03  & 2.21  \\
          & E8M7 & 1.36 & 1.60 & 1.53  & 1.71  & 1.73 
\end{tabular}
\label{tab:epsilons}
\end{table}

\begin{table}[ht]
\captionof{table}{
We show the loss landscape sharpness of a Llama 7B model initially trained with E8M3 precision for 6,000 training steps that was then trained with standard BF16. We can see that the sharpness indicator decreases in value with more training on BF16, indicating that it is capturing the increased stability of training that comes with BF16 over E8M3.
}
\begin{tabular}{rc}
\multicolumn{1}{r}{Train Step} & \multicolumn{1}{c}{Sharpness} \\ \hline
7K                              & 1.35                           \\
8K                              & 1.14                           \\
9K                              & 0.98                           \\
10K                             & 0.90                           \\
11K                             & 0.87                           \\
12K                             & 0.77                           \\
13K                             & 0.71                           \\
14K                             & 0.63                           \\
15K                             & 0.60                           \\
16K                             & 0.59                           \\
17K                             & 0.57                           \\
18K                             & 0.50                           \\
19K                             & 0.48                           \\
20K                             & 0.48                           \\
21K                             & 0.46                           \\
22K                             & 0.44                           \\
23K                             & 0.40                           \\
24K                             & 0.40                           \\
25K                             & 0.37                           \\
26K                             & 0.34                           \\
27K                             & 0.34                           \\
28K                             & 0.34                           \\
29K                             & 0.33                           \\
30K                             & 0.35                           \\
31K                             & 0.32                           \\
32K                             & 0.32                           \\
33K                             & 0.30                           \\
34K                             & 0.29                           \\
35K                             & 0.29                           \\
36K                             & 0.28                           \\
37K                             & 0.27                           \\
38K                             & 0.26                           \\
39K                             & 0.27                           \\
40K                             & 0.27                           \\
41K                             & 0.25                           \\
42K                             & 0.23                           \\
43K                             & 0.23                           \\
44K                             & 0.24                          
\end{tabular}
\label{tab:sharpness}
\end{table}

\begin{figure}[ht]
\centering
\begin{minted}{python}
def backward(inputs, weight, output_gradient):
    masked_inputs = reduce_precision(inputs)
    masked_weight = reduce_precision(weight)
    masked_output_gradient = reduce_precision(output_gradient)
    inputs_gradient = F.linear(masked_inputs, masked_weight.T)
    weight_gradient = F.linear(masked_output_gradient.T, masked_weight.T)
    masked_inputs_gradient = reduce_precision(inputs_gradient)
    masked_weight_gradient = reduce_precision(weight_gradient)
    return masked_inputs_gradient, masked_weight_gradient
\end{minted}
\captionof{figure}{
PyTorch-like pseudocode for the reduced-precision backward pass.
}
\end{figure}

\end{document}